# Arbitrary Handwriting Image Style Transfer

Kai Yang, Xiaoman Liang and Huihuang Zhao

Abstract: This paper proposed a method to imitate handwriting style by style transfer. We proposed a neural network model based on conditional generative adversarial networks (cGAN) for handwriting style transfer. This paper improved the loss function on the basis of the GAN. Compared with other handwriting imitation methods, the handwriting style transfer's effect and efficiency have been significantly improved. The experiments showed that the shape of the generated Chinese characters is clear and the analysis of experimental data showed the Generative adversarial networks showed excellent performance in handwriting style transfer. The generated text image is closer to the real handwriting and achieved a better performance in term of handwriting imitation.

## 1    Introduction

Neural networks learned from imitating the signal transmission methods of human brain neurons. In recent years, neural networks and deep learning have received more and more attentions and made great achievements in many fields. Since 2016, deep learning has been applied to a new field called style transfer[4]. Style transfer inputting a content picture and a style picture and then generate a new one after the feature extraction and reconstruction of the neural network. The picture has the content structure of the content picture and the artistic texture and style of the style picture.

This kind of research has become a very popular research topic in artificial intelligence. Neural style transfer is widely used in many fields to solve a variety of problems, such as video stylization[14], text style transfer[13],texture synthesis[3],and super-resolution[15],however,the handwriting is one of the most important way of communication for human beings. Chinese characters have a history of thousands of years. In the latest official GBK standard, Chinese characters contain a total of 27,533 Chinese characters. Everyone has their own handwriting style,but it is a very arduous and tedious task for creating a personal font library of 27,533 Chinese characters.This paper proposed a method for Handwriting imitation using style transfer by the conditional generative adversarial network.[1][5]After Compared with the previous fonts transfer methods, GAN has achieved a good performance in fonts style transfer. However, because of the complexity of Chinese characters, the methods in previous studies of fonts style transfer did not perform well on Chinese characters. In this paper, we use L1 loss function based GAN to solve this problem . Compared with other Loss functions L1 can retain the local detail structure.

捕 捕 伤 伤 猿 猿 觉 觉

**Fig . 1.** Comparison of generated fonts and source fonts,the left handwriting style is the target fonts.

## 2  Related work

The neural style transfer has achieved excellent results in various fields, what is impressive is that neural networks often perform beyond people's expectations in various fields.Many research institutions and laboratories have carried out their extensive and in-depth research on neural style transfer[4].Among them, the Texture style transfer [9][13]is one of the most popular research topics.

The term Generative Adversarial Network(GAN) was first proposed in 2014 by Ian Goodfellow of the University of Montreal and his colleagues[16]. The generative adversarial network can be regarded as a generative model which learns the probability distribution of these samples by learning a large number of data samples from the encoder. Due to the excellent learning ability and generation ability of the generation adversarial network. Since its birth in 2014,the GAN has been developed to many generation adversarial networks. Among these networks, the well-known networks including the Conditional Generative Adversarial Networks (cGAN)[1][5], which was proposed in the same year in 2014. Compared with the original GAN network,the cGAN adding conditional constraints to the generator and discriminator,and it can be seen as an improvement from the unsupervised learning generation adversarial network to the supervised learning generation adversarial network that established a solid basis for subsequent development and further research. In 2015, because researchers want to see what the generative network has learned to visualize the representation information, deep convolutional generative confrontation networks(DCGANs)[11] are proposed to clarify the representation information of the generated images.

The purpose of this paper is to find a new method to apply to font style transfer. After comparison with some common methods, we found that cGAN is the most suitable for font style transfer. Recently,some researchers like Radford A et.al[11] Fogel S et.al.[12]also proposed some methods for handwriting transfer.But it is not suitable for Chinese characters handwriting transfer.

## 3  Architecture

We take consideration of the encoder and decoder structure in CGAN[6],as a result our generator is composed of U-net[10] network structure, and our discriminator uses a convolutional PatchGAN [1]classifier.First of all, the input of the network is the image generated by the two fonts obtained by preprocessing and the random noise image generated by the generator.The U-net network of the generator extracts the feature map of the text image through the convolutional layer and the maxpooling layer of the encoder, and decoder expands the feature map upwards through deconvolution.

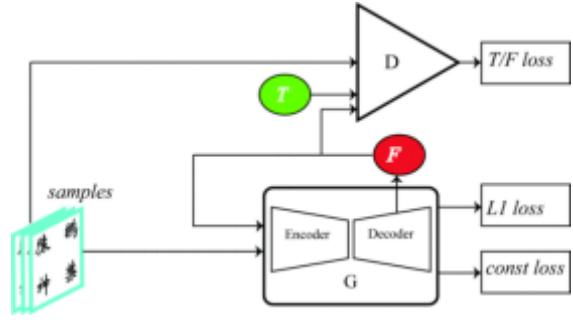

**Fig .2.** The GAN network structure we used, G means the generator and D means the discriminator.

The real target font image is input to the discriminator, and the random noise image generated by the generator from the source font is compared with the encoded real image inside the generator. The Constant loss function is used to optimize the generated random noise image to make it getting closer to the label picture.

$$L_{constant} = \sum_{i=1}^{n} \frac{1}{M} \left[ T^a_{i(w,h,c)} - F^a_{i(w,h,c)} \right]^2 \quad (1)$$

We not only need to constrain the loss between the image generated by the generator and the label image, but also need to constrain the loss between the generated image and the Source image, so we set up the L1 loss function to monitor the quality of the generated image. And we used the Tvloss loss function in the generator to eliminate blur noise caused by different fonts or scribbled handwriting.

$$L_1 = \sum_{i=1}^{n} \frac{1}{n} \left| T^b_{i(w,h,c)} - F^b_{i(w,h,c)} \right| \quad (2)$$

In GAN, one of the most important thing is that the image generated by the generator must fool the discriminator, so I. Goodfellow et.al [11] proposed cheat loss to maximize the probability of the generated image fooling the discriminator.

$$L_{cheat} = \frac{1}{n} \sum_{i=1}^{n} -\left[ y_i \cdot \log(p_i) + (1-y_i) \cdot \log(1-p_i) \right] \quad (3)$$

Where is $y_i$ represent the label of sample $i$ and $p_i$ represent the probability that the sample $i$ is predicted to be positive.

The discriminator must also need to be more excellent to judge the difference between the picture generated by the generator and the real handwritten picture. The competition between the generator and the discriminator makes the generated pictures more excellent. We continuously train the model by saving the parameters obtained from the training, finally we will get the neural network that can generate other text based on the handwriting images.

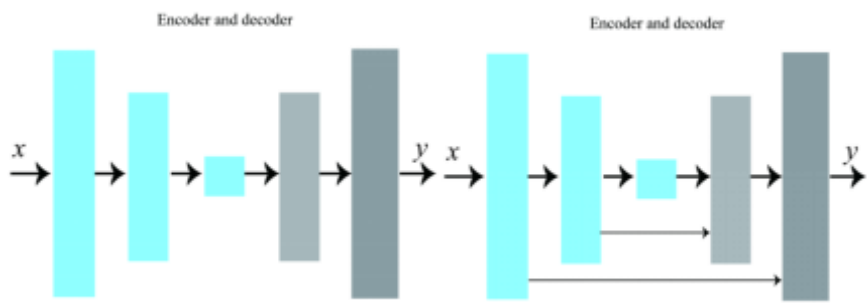

**Fig .3.**The encoder and decoder in the GAN neural network adopt the U-net structure to connect the encoder and decoder of the same layer through skip connection

## 4    Experiment

We hope that the cGAN model we trained can generalize and generate fonts of various complex styles. Because of the particularity of handwriting, it is difficult to imitate any person's handwriting style. Therefore, during training, we used a variety of different fonts for training, in order to allow the model to learn as many Chinese character styles as possible.

**Experiment environment** All the experiments are on a single Nvidia GTX1080 GPU and an Inter i7 3.7GHz CPU with Ubuntu16.04,which is efficient for our tasks.

**Running time** It takes about 36 hours on a single Nvidia GTX1080 GPU to train the model over 40 epochs, which may include about 32,000 examples.

### 4.1    Data set

Training this GAN neural network requires a lot of pictures of Chinese characters. How to find so many complete Chinese character libraries is a problem.At present, there is no outstanding public data set on the research topic of Chinese character handwriting imitation.

One method is to input a large amount of handwriting into the font generation software to obtain the required font images. Another way is purchase the official copyright of the font to obtain the required font library. Generally, this work of preparing a data set is huge.

In our work, the data set we used to train the cGAN network is the FangZheng Chinese font library. The font input to the network should have as many font libraries as possible. If the input source font and the trained label font cannot match a certain Chinese character, it is likely to cause the model to crash. It is recommended to save the parameters before each training of the network.In addition,preprocessing is necessary to pickle data into binary and save in memory during training.

罣　　　惡　　　誡　　　頴

**Fig. 4.**Some inappropriate samples to pickle data into binary,which will cause the failure to the model.

為人民服務
毛泽东

**Fig .5.**The handwriting calligraphy of the great Chinese leader Mao Zedong

### 4.2　Experiment result

Our experimental goal is to train a network model that can imitate human handwriting. After training about 20 fonts, we use some famous people's handwriting images as input to the GAN network for inference,Figure 5 and the results obtained are better than expected.

**Fig. 6.**The result that using Mao's handwriting style image as the input,which is learned from Fangzheng Lishu font library.

Because handwriting style is complicated, and the spacing between characters is difficult to divide into individual fonts, this will affects the effect of the experiment to a certain extent, but our purpose is to use deep learning to imitate any handwriting style. So we found a more complex data set to show the effect.

**Fig. 7.**The result that using Fangzheng Kaiti characters as the source font.

When our experiments with similar structure fonts and handwriting fonts, it will often get better results with clear shapes and no blurring of noise.

**Fig. 8.**Some samples that using a similar fonts style and handwriting.

The reason for this different results may be the structure or style of the text is too large to make it difficult for the constant loss to converge to an ideal situation.

## 5     Conclusion

In this paper,we proposed a new handwriting imitation method through deep learning neural networks, and constrain the gap between the generated handwritten font and the source font through the conditional generation adversarial network cGAN.This is a new way to imitating handwriting style. We set up a generators and variety of the loss functions of the discriminator to constrain the thickness and length of the strokes of the handwritten font.Then we trained with a variety of different styles of Chinese characters. Our generator used the encoder and decoder in the U-net network, and both the encoder and decoder used the form of convolutional neural network. It can be determined through experiments that good results can be achieved when both the encoder and decoder of U-net are trained through L1 loss. Fortunately,we have achieved excellent experimental results.

## 6     Expectation

On the research topic of handwriting imitation, we have done much but not completely work. Handwriting imitation has many interesting directions that can be expanded. For example, if researchers in the future can train a model that can imitate the handwriting style of all kinds of languages fonts or imitate the handwriting style of all languages, that would be a great achievement. On the other hand, our network model has some flaws when training the data set, such as the size of the text, different size of text may also cause bad results. If the network structure can be optimized in the future work to improve this situation or solve it, then it will be a great progress for deep learning handwriting imitation or text style transfer.

## 7     ACKNOWLEDGEMENTS

This work was supported by National Natural Science Foundation of China (61772179),Hunan Provincial Natural Science Foundation of China(2020JJ4152), the Science and Technology Plan Project of Hunan Province(2016TP1020),Double First-Class University Project of Hunan Province(Xiangjiaotong [2018]469), Postgraduate Scientific Research Innovation Project of Hunan Province(CX20190998),Degree & Postgraduate Education Reform Project of Hunan Province (2019JGYB266,2020JGZD072),Industry University Research Innovation Foundation of Ministry of Education Science and Technology Development Center (2020QT09), Hengyang technology innovation guidance projects(Hengcaijiaozhi [2020]-67), Postgraduate Teaching Platform Project of Hunan Province(Xiangjiaotong [2019]370-321).